\documentclass[10pt,twocolumn,letterpaper]{article}
\usepackage[utf8]{inputenc}
\usepackage{cvpr}              % To produce the CAMERA-READY version
\usepackage{graphicx}
\usepackage{booktabs}
\usepackage{multirow}% 导入booktabs宏包

\usepackage[dvipsnames]{xcolor}

\definecolor{cvprblue}{rgb}{0.21,0.49,0.74}
\usepackage[pagebackref,breaklinks,colorlinks,citecolor=cvprblue]{hyperref}

\def\paperID{19} % *** Enter the Paper ID here
\def\confName{CVPR}
\def\confYear{2024}

\title{Self-Supervised Learning of Deviation in Latent Representation for Co-speech Gesture Video Generation}

\author{
~~Huan Yang$^{1}$~~~~~~~~~Jiahui Chen$^{1,2}$~~~~~~~~~~Chaofan Ding$^1$~~~~~~~~Runhua Shi$^1$~~~~~~~~Siyu Xiong$^{1}$
\\~~~~~~~~Qingqi Hong$^2$~~~~~~~~~~Xiaoqi Mo$^1$~~~~~~~~~~Xinhan Di$^1$\\
$^1$ AI Lab, Giant Network\\
$^2$ Xiamen Univeristy\\
}

\begin{document}
\maketitle

\begin{abstract}
Gestures are pivotal in enhancing co-speech communication, while recent works have mostly focused on point-level motion transformation or fully supervised motion representations through data-driven approaches, we explore the representation of gestures in co-speech, with a focus on self-supervised representation and pixel-level motion deviation, utilizing a diffusion model which incorporates latent motion features. Our approach leverages self-supervised deviation in latent representation to facilitate hand gestures generation, which are crucial for generating realistic gesture videos. Results of our first experiment demonstrate that our method enhances the quality of generated videos, with an improvement from 2.7 to 4.5\% for FGD, DIV and FVD, and 8.1\% for PSNR, 2.5\% for SSIM over the current state-of-the-art methods. 
\end{abstract}%0.25-Page    
\section{Introduction}
\label{sec:intro}
Co-speech gestures, a fundamental aspect of human communication~\cite{tversky2007communicating}, convey information in tandem with speech. These gestures effectively transmit social cues~\cite{NGM2015Gesture}, including personality, emotion, and subtext. However, current conditional video generation methods~\cite{ruan2022mmdiffusion,Wang_2024_CVPR,Yan_2021,ho2022imagen,ceylan2023pix2video} often produce motion videos that lack realism and fine-grained details. Additionally, these networks typically require substantial computational resources~\cite{Chaitanya_2020,Yoon_2020,Ginosar_2019} and full supervision of motion annotation~\cite{corona2024vlogger} with high labour cost.

To address these challenges, we propose a novel approach for generating co-speech gesture videos. First, a deviation module is proposed to generate latent representation of both foreground and background. This deviation module is consisted of latent deviation extractor, a warping calculator and a latent deviation decoder. Second, a corresponding self-supervised learning strategy is proposed to achieve the latent representation of deviation for high-quality video generation. In the evaluation of our first experiment, the results demonstrate high-quality co-speech video generation and outperform the state-of-the-art models in the evaluation of video quality.

\begin{figure}
    \centering
    \includegraphics[width=0.50\textwidth]{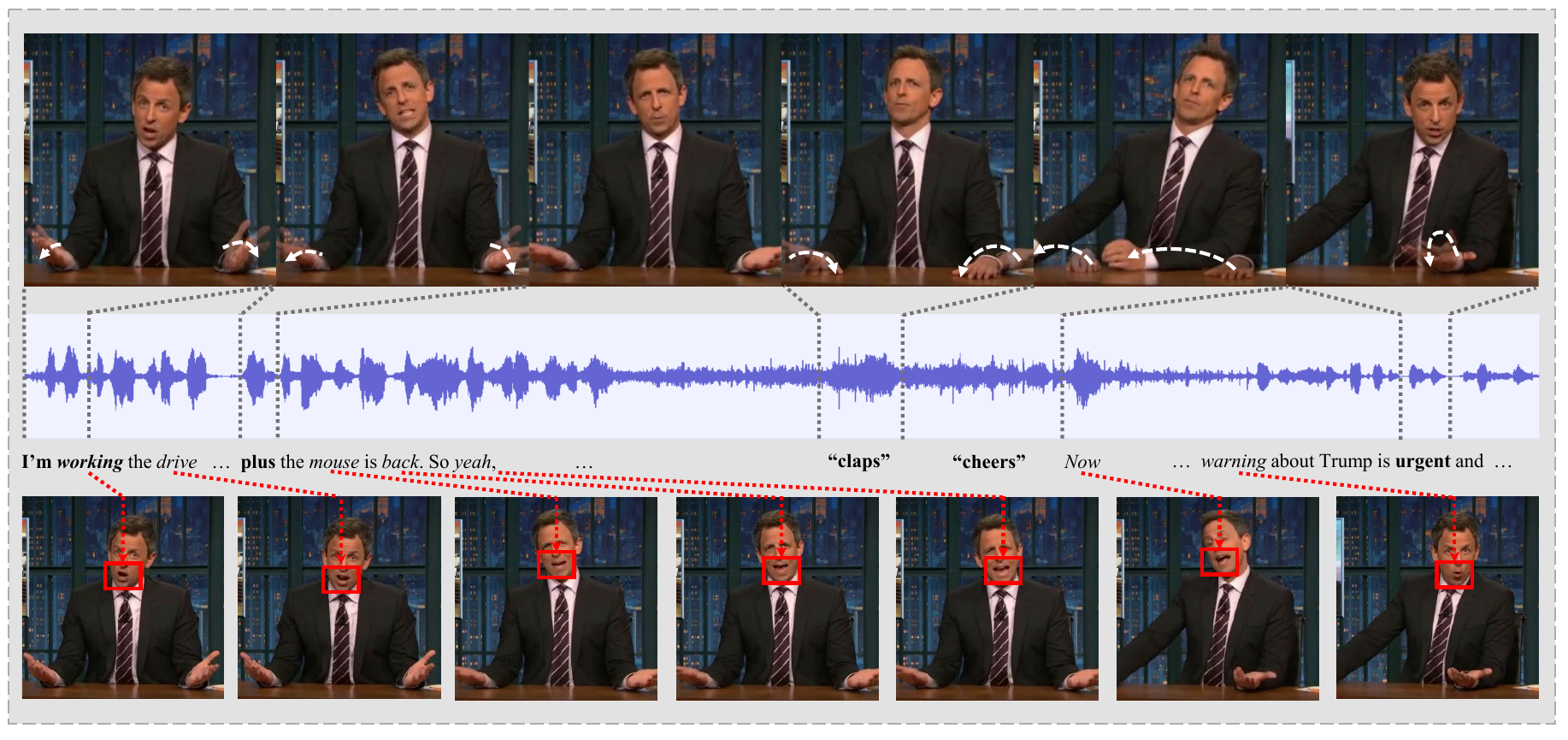}
    \caption{Examples of our generated gesture videos. White dashed arrows indicate gestures corresponding to bold words. The red dotted boxes indicate the mouth shapes corresponding to the italicized words.}
    \label{fig:fig1}
\end{figure}
%-------------------------------------------------------------------------

%0.5-Page
\section{Method}
\begin{figure}[ht!]
    \includegraphics[width=1.0\linewidth]{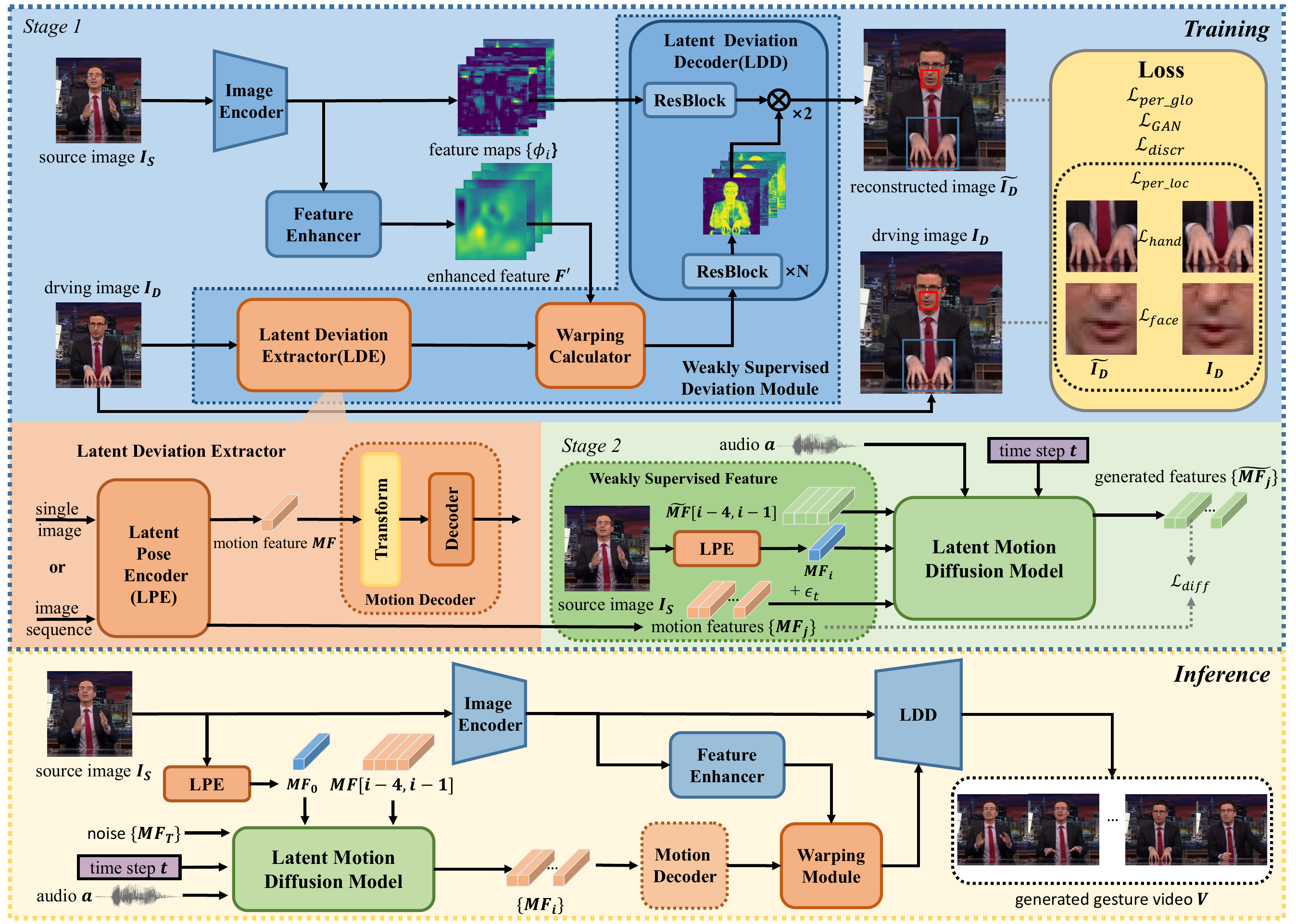}
    \caption{Co-speech gesture video generation pipeline of our proposed method consists of three main components: 1) the latent deviation extractor (orange) 2) the latent deviation decoder (blue) 3) the latent motion diffusion (green).}
    \label{fig:overview}
\end{figure}

We propose a novel method for generating co-speech gesture videos, utilizing a self-supervised full scene deviation, produces co-speech gesture video $V$ (i.e., image sequence) that exhibit natural poses and synchronized movements. The generation process takes as input the speaker's speech audio $a$ and a source image $I_S$. An overview of our model is shown in Figure~\ref{fig:overview}. 

We structured the training process into two stages. In the first stage, a driving image $I_D$ and a source image $I_S$ are used to train the base model. In one aspect, the proposed latent deviation module consisting of latent deviation extractor, warping calculator and latent deviation decoder is trained under self-supervision. In another aspect, other modules in the base model is trained under full supervision. In the second stage, the self-supervised motion features, consisting of $MF_i$, $\widetilde{MF}_{[i-4,i-1]}$, and the noise-added motion feature sequence $\{MF_j\}$, are used to train the latent motion diffusion model. In the following parts, we will introduce the two stages of training part and the inference part in detail. 
\begin{figure}
    \centering
    \includegraphics[width=\linewidth]{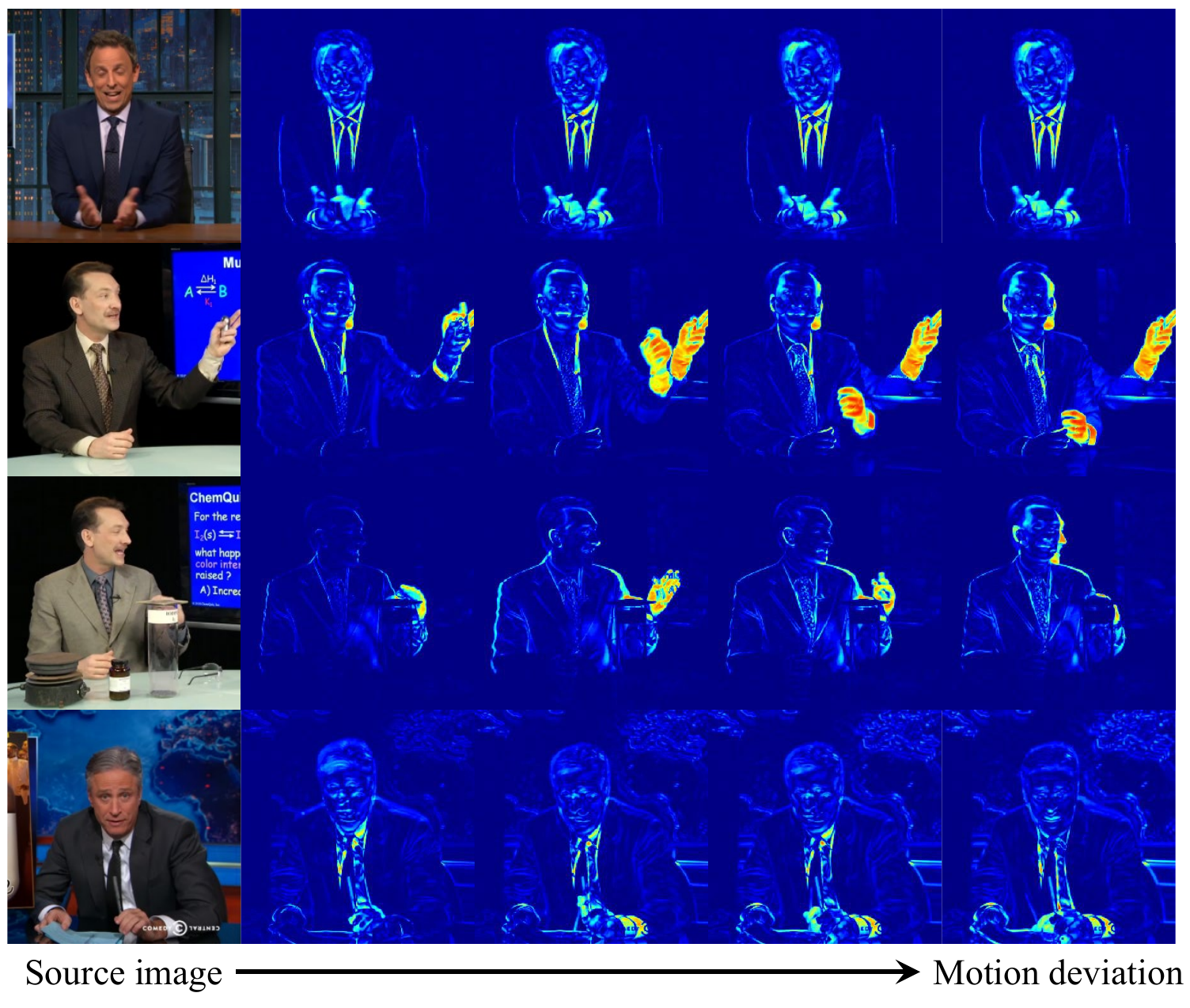}
    \caption{The deviation in latent representation.}
    \label{fig:deviation}
\end{figure}
\subsection{Stage 1: Base Model Learning}

\subsubsection{Image encode.} 
First, we input the source image $I_S \in \mathbb{R}^{H\times W \times 3}$ into the image encoder $\mathcal{E}$ to obtain the feature $F$ and feature maps $\{\phi_{i\in N}\}$, as the layer-by-layer feature maps provide additional background details for subsequent steps. 
\begin{equation}
    F, \{\phi_{i\in N}\} = \mathcal{E}(I_S).
\end{equation}

\subsubsection{Feature Enhancer.} 
Directly inputting shallow spatial features $F$ into the warping calculator increases the training difficulty of LME and makes it more likely to decode blurred images with artifacts after applying optical flow transformations. Therefore, we use a feature enhancer to map the shallow spatial features to a higher-level space, achieving better results. The equation is as follows
\begin{equation}
    F' = \frac{F - \bar{F}}{\sqrt{\sigma^2+\epsilon}}\gamma + \beta,
\end{equation} where $\bar{F}$ and $\sigma$ represent the mean and standard deviation of the features, while $\gamma$ is the scaling factor, and $\beta$ is the bias.

\subsubsection{Self-Supervised Deviation Module.} 
We proposed a self-supervised deviation module which is consisted of three parts, latent deviation extractor, warping calculator and latent deviation decoder. The calculation is represented in the following.

\paragraph{Latent Deviation Extractor.} To generate natural motion-transformed images, we first design latent pose estimation module (LPE) to extract a latent motion feature $MF \in \mathbb{R}^{1 \times K}$. This latent motion feature then undergoes a nonlinear pose transformation operation before being decoded. The compact representation of pose transformations within the latent space enables the generation of concise optical flow, which effectively drives the image.
\begin{equation}
    \mathbf{V} = \psi(\mathcal{T}(LPE(I))).
\end{equation}

\paragraph{Warping Calculator.} To effectively integrate motion information into the source image, we first input the source image $I_S$ into the encoder, enhance the feature maps, and then warp $\mathcal{W}(\cdot)$ each enhanced feature using the rotation $\mathbf{R}$ and translation $\mathbf{T}$ matrices of optical flow $\mathbf{V}$ to obtain the deformed features $F'_\mathcal{W}$. This feature enhancement process amplifies key features while suppressing background noise, leading to a clearer representation of critical information in the image. The warping formula is as follows:
\begin{equation}
    F'_\mathcal{W} = \mathcal{W}(\mathbf{R}, \mathbf{T}, F').
\end{equation}

\paragraph{Latent Deviation Decoder.} Due to the occlusions and mis-alignments between $I_S$ and $I_D$, directly decoding after the warping operation often fails to achieve effective image reconstruction. Inspired by~\cite{zhao2022thin,corona2024vlogger}, we add an full scene deviation $\delta_F$ into the decoder during the decoding process, which improves the accuracy of the reconstructed image $\widetilde I_D$. The full scene deviation formula is as follows:
\begin{equation}
    \delta_F = L \frac{1}{1 + e^{-(wF'_\mathcal{W} + b)}},
\end{equation} which is a variant of the sigmoid function. 
\begin{figure}
    \includegraphics[width=\linewidth]{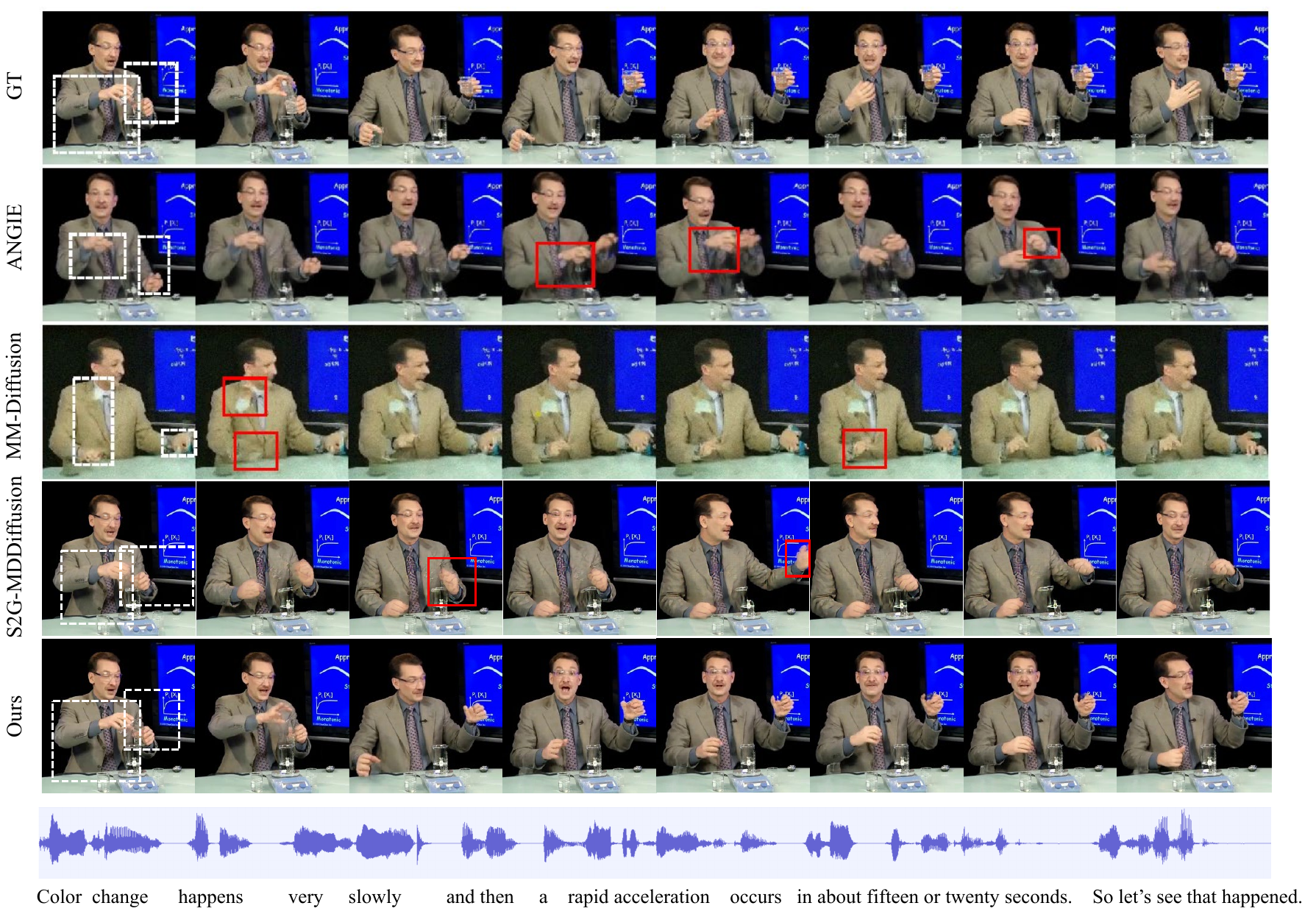}
    \caption{Visual comparison with SOTAs. Our method generates gestures with more extensive accurate motions (dashed boxes), matching audio and semantics. Red boxes indicate unrealistic gestures generated by ANGIE~\cite{liu2022audio}, MM-Diffusion~\cite{ruan2022mmdiffusion} and S2G-MDDiffusion~\cite{he2024co}.}
    \label{fig:comparisons1}
\end{figure}
Next, we interpolate and decode the deviation $\delta_F$ together with the feature $F$. This approach allows the motion features to be smoothly interpolated onto the source image features, resulting in more natural outcomes. The decoding process is as follows:
\begin{equation}
    \mathbf{z} = \delta_F F + (1 - \delta_F) U(F), 
\end{equation} where $U(\cdot)$ denotes image decoder. Moreover, we design a nonlinear activation function that can adjust the impact of negative input values. By tuning the parameter $c_{\lambda}$, we can control the contribution of negative input values to the output.
\begin{equation}
    \mathbf{a} = max(0, \mathbf{z}) + c_{\lambda} min(0, \mathbf{z}).
\end{equation}

\paragraph{Training.} 
In the training process of co-speech gesture video generation, unlike vlogger~\cite{corona2024vlogger} which applies a large-scale of annotation data for supervised dense representation of motion, we directly apply image reconstruction loss without dense/sparse supervised annotation of motion~\cite{bhat2023self}. Following~\cite{Siarohin_2019_NeurIPS, Siarohin2021motion,bhat2023self}, we utilize a pre-trained VGG-19~\cite{johnson2016perceptual} network to calculate the reconstruction global loss $\mathcal{L}_{per\_glo}$ between the reconstructed image $I_D$ and the generated image $\widetilde I_D$ across multiple resolutions:
\begin{equation}
    \mathcal{L}_{per\_glo} = \sum_{j} \sum_{i} c_i |V_{i}(I_{Dj}) -V_{i}(\widetilde I_{Dj})|,
\end{equation}
where $V_i$ is the $i$-th layer of the pre-trained VGG-19 network, and $j$ represents that the image is downsampled $j$ times. Additionally, we calculated the local loss $\mathcal{L}_{per\_loc}$ of the reconstructed image, which consists of hand loss $\mathcal{L}_{hand}$ and face loss $\mathcal{L}_{face}$, based on the VGG-16 model:
\begin{equation}
    \mathcal{L}_{per} = \lambda_{per\_glo}\mathcal{L}_{per\_glo} + \lambda_{per\_loc}\mathcal{L}_{per\_loc}.
\end{equation}

To ensure realism, we employed a patch-based discriminator~\cite{quan2022image} and trained it using the GAN discriminator loss $\mathcal{L}_{discr}$ for adversarial training. Both the ground truth images and the generated fine images are converted into feature maps, where each element is classified as real or fake. Therefore, the final loss is the following:
\begin{equation}
    \mathcal{L}_1 = \lambda_{per}\mathcal{L}_{per} + \lambda_{GAN}\mathcal{L}_{GAN} + \lambda_{discr}\mathcal{L}_{discr}.
\end{equation}

\subsection{Stage 2: Latent Motion Diffusion}

In this stage, we adopted the latent motion diffusion model proposed by~\cite{he2024co} and made modifications to it. Our diffusion model takes five inputs: the time step $t$, audio $a$, the noisy motion feature sequence $\{MF_{j\in M}\}$ , the motion feature ${MF}_i$ of the source image, and the predicted motion features $\widetilde {MF}_{[i-4, i-1]}$ from the previous four frames. The model then predicts a clean motion feature sequence $\{\widetilde {MF}_{i\in M}\}$ from noised $\{MF_{j\in M}\} + \epsilon_t$ and condition $c = (a, \widetilde {MF}_{[i-4, i-1]}, {MF}_i )$.
\begin{figure}
    \centering
    \includegraphics[width=\linewidth]{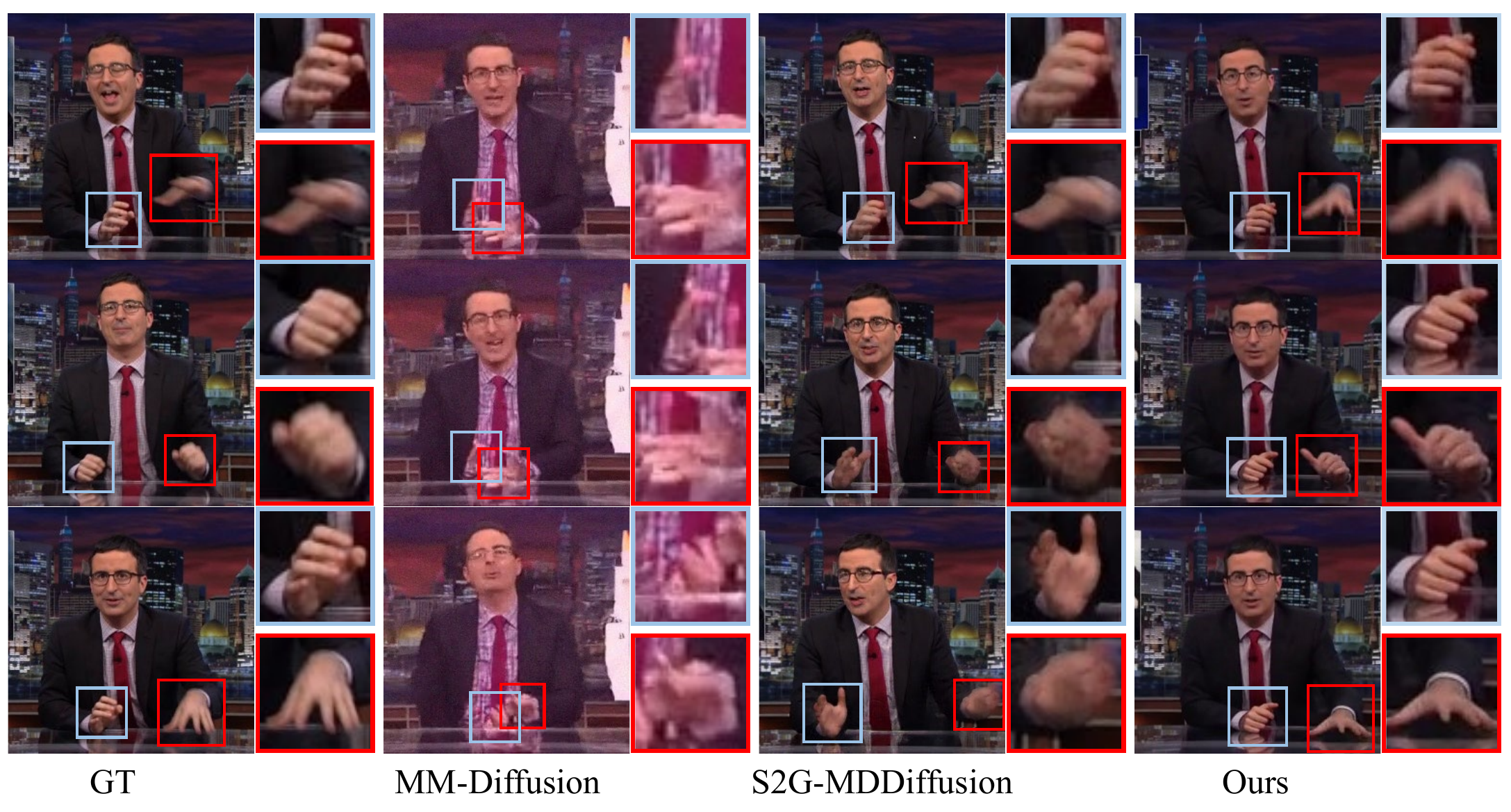}
    \caption{Visualization results of fine-grained hand variations. The gesture videos we generate are clearer, more reasonable, more diverse and more natural in the same frame.}
    \label{fig:comparisons2}
\end{figure}
\subsubsection{Feature Priors and Loss.} 
Both self-supervised deviation feature and fully-supervised feature are applied. We divide the loss into three components. The first is a motion feature loss calculated with MSE, which constrains the movements of the hands, lips, and head, promoting naturalness and coherence. Additionally, the loss includes implicit velocity and implicit acceleration losses~\cite{li2022cvpr} to prevent the overall motion from being too rapid. The final training loss is as follows:
\begin{equation}
    \mathcal{L}_{diff} = \mathcal{L}_{MF} + \lambda_{im\_vel}\mathcal{L}_{imp\_vel} + \lambda_{imp\_acc}\mathcal{L}_{imp\_acc}.
\end{equation}

\begin{table*}
    \centering
    \begin{tabular}{c c c c c c c c}
    \multirow{2}{*}{Name} & \multicolumn{3}{c}{Objective evaluation} & \multicolumn{4}{c}{Subjective evaluation} \\
    \cline{2-8}
    \multirow{2}{*}{} & FGD $\downarrow$ & Div. $\uparrow$ & FVD $\downarrow$ & Realness $\uparrow$ & Diversity $\uparrow$ & Synchrony $\uparrow$ & Overall quality $\uparrow$ \\
    \hline
    Ground Truth (GT) & 8.976 & 5.911  & 1852.86 & 4.70$\pm$ 0.07 & 4.45 $\pm$ 0.05 & 4.66 $\pm$ 0.08 & 4.70 $\pm$ 0.07 \\
    \hline
    ANGIE & 55.655 & 5.089  & 2965.29 & 2.01$\pm$0.07 & 2.38$\pm$0.07 & 2.11$\pm$0.07 & 1.98$\pm$0.08 \\
    MM-Diffusion & 41.626 & 5.189 & 2656.06 & 1.63$\pm$0.07 & 1.93$\pm$0.08 & 1.57$\pm$0.09 & 1.44$\pm$0.08 \\
    S2G-MDDiffusion & \underline{18.131} & \underline{5.632}  & \underline{2058.19} & \underline{2.78$\pm$0.07} & \underline{3.21$\pm$0.08} & \underline{3.32$\pm$0.08} & \underline{2.89$\pm$0.07} \\
    Ours & \textbf{17.324} & \textbf{5.788} & \textbf{1997.96} & \textbf{3.82$\pm$0.07} & \textbf{3.94$\pm$0.07} & \textbf{4.11$\pm$0.06} & \textbf{3.93$\pm$0.05} \\
    \hline
    \end{tabular}
    \caption{Quantitative results on test set. Bold indicates the best and underline indicates the second. For ANGIE~\cite{liu2022audio} and MM-Diffusion~\cite{ruan2022mmdiffusion}, we cited the results in the S2G-MDDiffusion~\cite{he2024co} paper. For S2G-MDDiffusion, we used the official open source code.}
    \label{tab:quantitative}
\end{table*}

\begin{table*}[ht!]
    \centering
    \begin{tabular}{c c c c c c c c c c}
        \multirow{2}{*}{Name} & \multicolumn{3}{c}{Hand gesture} & \multicolumn{3}{c}{Lip movement} &  \multicolumn{3}{c}{Full image}\\
        \cline{2-4} \cline{5-7} \cline{8-10}
        & PSNR $\uparrow$ & SSIM $\uparrow$ & LPIPS $\downarrow$ & PSNR $\uparrow$ & SSIM $\uparrow$ & LPIPS $\downarrow$ & PSNR $\uparrow$ & SSIM $\uparrow$ & LPIPS $\downarrow$ \\
        \hline
        S2G-MDDiffusion & 22.37 & 0.625 & 0.106 & 23.84 & 0.689 & 0.092 & 29.39 & 0.952 & 0.034 \\
        Ours & \textbf{23.91} & \textbf{0.756} & \textbf{0.054} & \textbf{29.10} & \textbf{0.882} & \textbf{0.038} & \textbf{31.79} & \textbf{0.976} & \textbf{0.018} \\
        \hline
    \end{tabular}
    \caption{Comparison of gestures and mouth movements based on common image quality metrics.}
    \label{tab:comparisons2}
\end{table*}

\begin{table*}[ht!]
    \centering
    % \resizebox{\textwidth}{15mm}{
    \begin{tabular}{c c c c c c c c}
    \multirow{2}{*}{Name} & \multicolumn{3}{c}{Objective evaluation} & \multicolumn{4}{c}{Subjective evaluation} \\
    \cline{2-8}
    \multirow{2}{*}{} & FGD $\downarrow$ & Div. $\uparrow$ & FVD $\downarrow$ & Realness $\uparrow$ & Diversity $\uparrow$ & Synchrony $\uparrow$ & Overall quality $\uparrow$ \\
    \hline
    w/o Dev. in LDD & \textbf{17.218} & 4.980  & \underline{2112.81} & 3.03 $\pm$ 0.08 & 3.62 $\pm$ 0.08 & 3.62 $\pm$ 0.07 & 3.29 $\pm$ 0.08 \\
    w/o $F'$ & \underline{17.284} & \underline{5.703}  & 2252.88 & 3.62 $\pm$ 0.06 & 3.81 $\pm$ 0.09 & 3.80 $\pm$ 0.08 & 3.67 $\pm$ 0.08 \\
    w/o Mo Dec. & 17.812 & 5.191 & 2201.66 & \underline{3.66 $\pm$ 0.07} & \underline{3.85 $\pm$ 0.09} & \underline{3.86 $\pm$ 0.06} & \underline{3.77 $\pm$ 0.06} \\
    \hline
    Ours & 17.324 & \textbf{5.788} & \textbf{1997.96}  & \textbf{3.87$\pm$0.07} & \textbf{3.86$\pm$0.08} & \textbf{4.08$\pm$0.07} & \textbf{3.86$\pm$0.08} \\
    \hline
    \end{tabular}
    % }
    \caption{Ablation study results. Bold indicates the best and underline indicates the second. 'w/o' is short for 'without'.}
    \label{tab:ablation}
\end{table*}
%0.75-Page
\section{Experiments and Results}
%\label{sec:rationale}
\subsection{Dataset and Evaluation metrics.}
Our experimental data comes from the PATS dataset~\cite{Chaitanya_2020}, following the standard process from S2G-MDDiffusion~\cite{he2024co}, we conduct the first training stage to generate the gesture videos and compare with the state-of0the-art models.

We employed 1) \textbf{Fréchet Gesture Distance (FGD)} 2) \textbf{Fréchet Video Distance (FVD)} and 3) \textbf{Diversity (Div.)} to assess quality of gesture videos~\cite{he2024co}. Besides, we also apply PSNR, SSIM and LIPS~\cite{corona2024vlogger} for the evaluation of generated hand gesture, lip movement and full scene. 

\subsection{Evaluation on Results(First Stage)}

We compared our method with: 1) gesture video generation methods ANGIE~\cite{liu2022audio} and S2G-MDDiffusion~\cite{he2024co}, and 2) MM-Diffusion~\cite{ruan2022mmdiffusion}.

In Figure~\ref{fig:comparisons1}, Our method(first stage) mitigate abnormal deformations in comparison with the SOTA models. In Figure~\ref{fig:comparisons2}, we also compare generation of hand gesture, facial expression and lip movements, our method also improves the quality of these parts. In detail, the proposed method mitigates jitter, blurred region(hands/faces/lips) and other abnormal deformations.

The quantitative results are shown in Table~\ref{tab:quantitative} and~\ref{tab:comparisons2}. Our method outperforms existing approaches in terms of FGD, Diversity, and FVD metrics, PSNR, SSIM and LIPS. Compared to S2G-MDDiffusion, our method achieved a 4.45\% and 2.93\% reduction in FGD and FVD, respectively, and a 2.77\% increase in DIV. In terms of image quality, our method outperforms S2G-MDDiffusion across the metrics for hands, lips, and the entire image for PSNR(6.88\%-22.06\%) and for SSIM(2.52\%-28.01\%). 
 
\subsection{User Study and Ablation Study}
We conducted a user study inspired by S2G-MDDiffusion~\cite{he2024co}. Twenty participants were invited to provide Mean Opinion Scores (MOS) across four dimensions, including 1) \textbf{Realness}, 2) \textbf{Diversity} of gesture, 3) \textbf{Synchrony} between speech and gestures, and 4) \textbf{Overall quality}. Besides, We examined the effectiveness of the following components: 1) the deviation in the LDD, 2) the enhanced feature, and 3) the motion decoder in Table~\ref{tab:ablation}.

\section{Discussion}
We proposed the self-supervised learning of deviation in co-speech gesture video generation. The results(first stage) demonstrates improvement of the quality of gesture videos. We are conducting the second stage of the experiment.

% \begin{table}[ht!]
%     \centering
%     \begin{tabular}{c c c c}
%         Name & PSNR & SSIM & LPIPS \\
%         \hline
%         S2G-MDDiffusion & 23.84 & 0.689 & 0.092 \\
%         Ours & 29.10 & 0.882 & 0.038 \\
%         \hline
%     \end{tabular}
%     \caption{Caption}
%     \label{tab:comparisons3}
% \end{table}
%0.75-Page
% \input{sec/5_Experiment}%0.75-Page
\bibliographystyle{plain}
\bibliography{main}


\end{document}